\documentclass[letterpaper, 10 pt, conference]{ieeeconf}  % Comment this line out if you need a4paper

\IEEEoverridecommandlockouts                              % This command is only needed if 
\usepackage{graphicx}
\usepackage{hyperref}
\title{\LARGE \bf
HyPerNav: Hybrid Perception for Object-Oriented Navigation in Unknown Environment
}

\author{Zecheng Yin$^{1,3}$, Hao Zhao$^{2}$, Zhen Li$^{1,3}$% <-this % stops a space
\thanks{$^{1}$ Future Network of Intelligence Institute(Shenzhen),
        Shenzhen, China
        {\tt\small yinzecheng.cuhk@gmail.com}}%
\thanks{$^{2}$Tsinghua University, Beijing, China
        {\tt\small }}%
\thanks{$^{3}$The Chinese University of Hongkong(Shenzhen), Shenzhen, China
        {\tt\small lizhen@cuhk.edu.cn}}%
}

\begin{document}

\maketitle
\thispagestyle{empty}
\pagestyle{empty}

%%%%%%%%%%%%%%%%%%%%%%%%%%%%%%%%%%%%%%%%%%%%%%%%%%%%%%%%%%%%%%%%%%%%%%%%%%%%%%%%
\begin{abstract}

Objective-oriented navigation(ObjNav) enables robot to navigate to target object directly and autonomously in an unknown environment. Effective perception in navigation in unknown environment is critical for autonomous robots. While egocentric observations from RGB-D sensors provide abundant local information, real-time top-down maps offer valuable global context for ObjNav. Nevertheless, the majority of existing studies focus on a single source, seldom integrating these two complementary perceptual modalities, despite the fact that humans naturally attend to both. With the rapid advancement of Vision-Language Models(VLMs), we propose Hybrid Perception Navigation (HyPerNav), leveraging VLMs' strong reasoning and vision-language understanding capabilities to jointly perceive both local and global information to enhance the effectiveness and intelligence of navigation in unknown environments. In both massive simulation evaluation and real-world validation, our methods achieved state-of-the-art performance against popular baselines. Benefiting from hybrid perception approach, our method captures richer cues and finds the objects more effectively, by simultaneously leveraging information understanding from egocentric observations and the top-down map. Our ablation study further proved that either of the hybrid perception contributes to the navigation performance. The code and datasets are publicly available.% at anonymous repository: \url{https://anonymous.4open.science/r/HyPerNav-88D1}.

\end{abstract}

%%%%%%%%%%%%%%%%%%%%%%%%%%%%%%%%%%%%%%%%%%%%%%%%%%%%%%%%%%%%%%%%%%%%%%%%%%%%%%%%
\section{Introduction}
Navigating to target objective from human language is a key ability for fully autonomous robots. Robots can accomplish amazing human language following with assisting spatial information in instructions or map such as Vison Language Navigation in Continuous Environment (VLN-CE)\cite{vlnce,qiao2025opennav,groundnav,nava3}. However, the environment prior from instructions or pre-constructed map limit the autonomy and generalization of navigation. 

We focus on Object-oriented navigation (ObjNav). Unlike the above, ObjNav requires the robot to directly moves to target in unknown environments given only the object goal text without any prior to the environment, thus demanding more on spatial perception, clue discovery and common sense reasoning in the unexplored environment. The evaluation of such task is the navigation success rate (SR) and Success weighted by Path Length(SPL)\cite{embodied_eval}. The success rate implies the possibility the robot successfully navigates to target objects and the SPL implies how close the robot trajectory to the shortest path to the objective.

To address the Object Navigation (ObjNav) problem, various approaches have been proposed, including classical exploration methods, learning-based methods, and training-free methods.  Classical approaches, such as frontier-based exploration \cite{frontier}, provide a basic baseline: the robot systematically traverses unknown areas while using a target detector to identify the goal. Once the target is detected during exploration, the robot navigates to it—examples include \cite{chang2023goatthing}. However, this strategy often requires extensive exploration time and is prone to failure in large or complex environments.

Learning-based methods employ reinforcement learning (RL) to improve the effectiveness of navigation. These approaches train an end-to-end navigation policy via dataset before deploying it in evaluation scenarios, as exemplified by \cite{chaplot2020semantic}. Recent learning-based models\cite{navidiffusor,nomad} also utilize diffusion to predict possible trajectories in egocentric observations to objectives. and its follow-up works. However, such methods face significant challenges in data acquisition and generalization, as they require large amounts of diverse training data—a requirement that currently limits their scalability and real-world applicability. Subsequent work\cite{chang2023goatthing,khanna2024goatbench}, extends the goal modality from text to images and incorporates lifelong memory mechanisms, but does not fundamentally address the core challenges in navigation policy learning.

Training-free methods have gained increasing popularity due to the rapid advancements in VLMs and similarity matching models\cite{blip2,clip}.  Some approaches leverage the commonsense reasoning capabilities of VLM to rank the frontiers for navigation \cite{kuang2024openfmnav}. Some\cite{vlmnav,navvlm} combine the egocentric observation and objective text prompt as a Vision Question Answering(VQA) problem for VLM to guide the navigation. \cite{ji2025dynavlmzeroshotvisionlanguagenavigation} follows \cite{vlmnav} in navigation guidance and adds Retrieval-Augmented Generation(RAG) for object memory management. Another line of work \cite{vlfm} employs the BLIP model \cite{blip2} to perform text-image matching in egocentric observation, enabling the robot to select the most promising frontier from local observations for intelligent exploration. Similarly, \cite{gmap} applies the same similarity-based strategy but combined all observations onto a global map, allowing for holistic goal localization.  Additionally, some efforts focus on training end-to-end large VLMs for embodied navigation \cite{embodied_generalist,vebrain,robotron}; however, these are currently limited by the scarcity of large-scale datasets and the computational constraints of resource-limited robotic platforms.

The mentioned works accomplish remarkable progress, however, they primarily rely on either local, egocentric perception \cite{navidiffusor,vlfm,navvlm,vlmnav,ji2025dynavlmzeroshotvisionlanguagenavigation} or global information derived from top-down navigation maps \cite{gmap,chaplot2020semantic,traj_diffusion}. Crucially, these approaches fail to organically integrate these two complementary sources of perceptual information. Moreover, several of them suffer from limited generalization capabilities \cite{chaplot2020semantic,nomad,vebrain}. 

To address these limitations, we propose Hybrid Perception Navigation (HyPerNav), with the following contributions:
\begin{itemize}
    \item We propose a  simple yet effective training-free VLM-based method that leverages hybrid perception, combining local and global perception for effective object-oriented navigation.
    \item We conducted simulation experiments with quantitative results and real-world validation demonstrating state-of-the-art performance.
    % \item Our method is fast and reliable on ObjNav compared to previous works. 
    \item Navigation intelligence scales with advancing VLMs still far from their potential. Our publicly released code provides a solid platform for future research.
\end{itemize}

\section{Related Works}

RGB-D egocentric perception serves as the fundamental sensory input for embodied agents, providing detailed observations of nearby objects, providing cues for goal-directed navigation, enabling collision avoidance, and facilitating the gradual construction of a global navigation map. 

\cite{chang2023goatthing} detects goal objects using \cite{detic} during classical frontier exploration\cite{frontier} and memorizes them along the trajectory. However, its ObjNav strategy relies solely on exhaustive exploration and object detector performance, resulting in inefficient and often ineffective navigation.

This egocentric, detection-based exploration paradigm has been advanced by \cite{vlfm,kuang2024openfmnav,ovon}, which leverage text-image similarity models \cite{blip2,siglip} or VLMs to rank frontier views by their relevance to the goal, guiding more intelligent exploration. For example, VLFM uses BLIP2 \cite{blip2} to compute similarity between a frontier image and a prompt like "there seems to be a bed ahead", selecting the most promising direction. This idea also extends to global perception: instead of comparing frontiers, \cite{gmap} projects similarity scores onto a top-down map obstacles to guide exploration at the global level.

Learning-based methods have also been proposed. \cite{nomad,navidiffusor} use diffusion models to generate real-time, egocentric future trajectories in local observation space. In contrast, \cite{diffusion_as_reason,traj_diffusion} predict target locations or trajectories directly in the top-down map. However, these approaches typically focus on either local or global observations during training and often suffer from slow inference due to the nature of diffusion sampling. 

End-to-end embodied foundation models \cite{vebrain,embodied_generalist} have emerged, but their development is constrained by the scarcity of diverse, high-quality training data required for robust cross-environment navigation. 

VLM-guidance-based navigation methods are emerging as a promising direction toward more intelligent object navigation. VLMs take image and language inputs and generate text outputs grounded in commonsense reasoning, serving as a "brain" for agents to act more intelligently during navigation. \cite{vlmnav,navvlm} take each step as a Vision Question Answering (VQA) task, feeding the current observation and a goal-related text prompt into the VLM to answer a set of predefined directional answers. For example, \cite{vlmnav,ji2025dynavlmzeroshotvisionlanguagenavigation} presents depth-derived candidate positions marked with IDs for the VLM to select from based on visual and prompt context, while \cite{navvlm} directly outputs the promising area in egocentric observation and then moves towards it.

With designed prompts, these methods can capture cues from perception during navigation and achieved remarkable performance. However, they are typically limited to a single perceptual modality, as they use either egocentric local views or global top-down maps, failing to integrate complementary information, which ultimately affects their navigation capabilities. This limitation is especially evident in methods relying solely on local perception, which often fail to escape corners, even when actions like ``turn around" are explicitly provided.

To address this, we propose HyPerNav, a hybrid perception method that combines both local and global views. Our design is inspired by human intuition: when observing a robot navigate, humans naturally attend to both immediate surroundings and the broader spatial context. HyPerNav leverages local perception to detect goal-relevant cues and uses global structure for efficient, long-range planning. It is training-free, easy to implement, and reproducible. HyPerNav is also faster than previous VLM-assisted baselines. These advantages are illustrated later through large-scale simulation evaluations and real-world experiments validations.

% --- despite being able to notice abundant visual cues and objects, they fail to integrate global contextual information, For example, they frequently struggle to escape from corners, even when the option of "turn around" is explicitly provided to the VLM.

\section{Methodology}
\begin{figure*}
    \centering
    \includegraphics[width=1\linewidth]{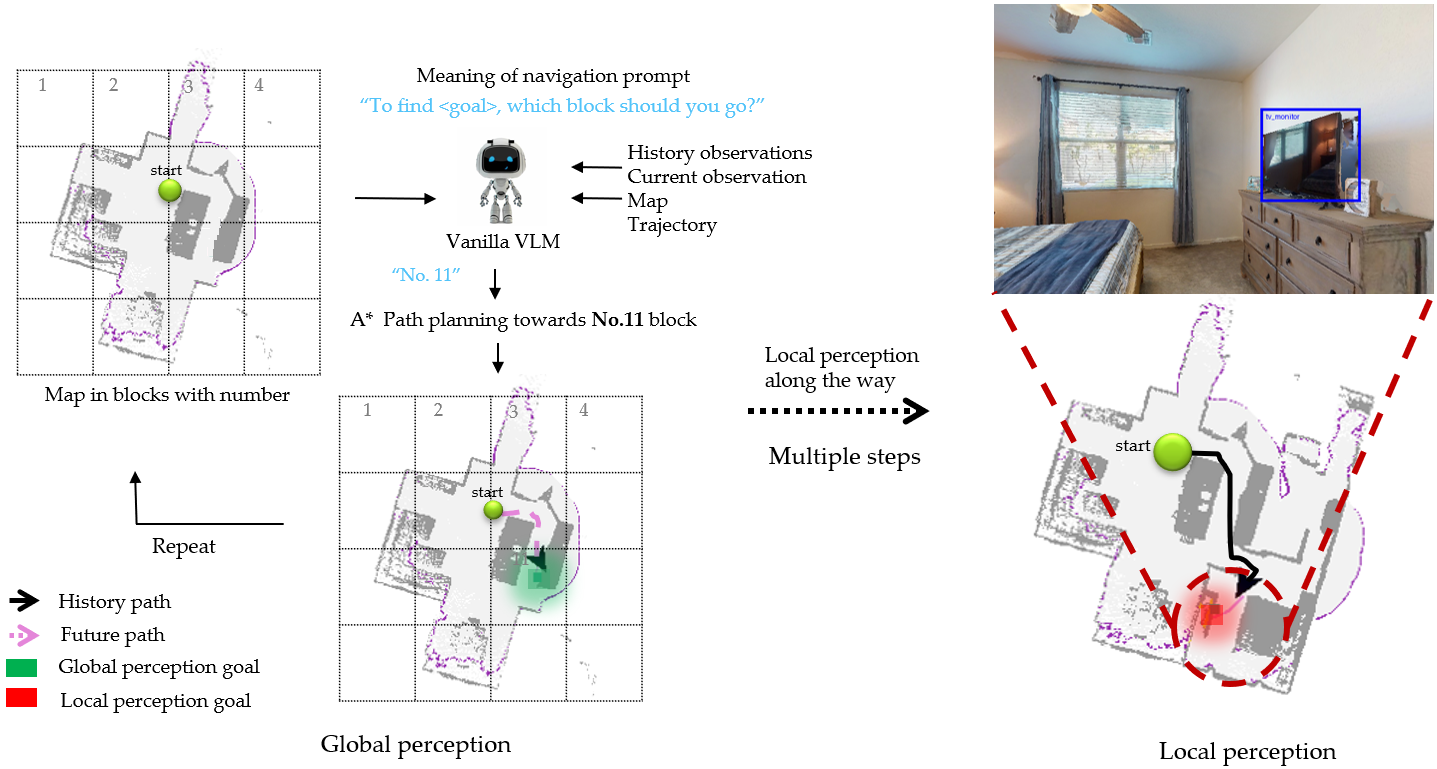}
    \caption{HyPerNav overall process. The left side illustrates the robot explores promising area based on global perception guidance via VLM answers. The right side explains local perception keeps finding the objects along the way until reaching the objectives. The grid map and its corresponding ID is dynamically updated based on explored area. Global perception provide potential target areas area and local perception provides final goal position. The path towards these positions is planned by path planning module based on A star algorithm.}
    \label{fig:process}
\end{figure*}

HyPerNav contains three main modules: local perception for egocentric observation cues, global perception for top-down map cues and path planning. The overall process visualization of our proposed work is in Figure \ref{fig:process}. Our method requires a robot equipped with an egocentric RGB-D sensor.

\subsection{Intuition}
The goal of our method is to navigate to a target object in an unknown environment. Local RGB observations provide rich information about the object's location and allow it to be precisely segmented from the scene. At the same time, the top-down map of the environment, initially built for path planning, also plays an important role in holistic spatial perception.
\begin{itemize}
    \item The top-down map contains rich structural information, where the shape of obstacles can provide useful navigation cues. For example, a "toilet" is more likely to be located in a small room, a "table" often appears as a large circular obstacle surrounded by smaller circles (chairs), and a partially explored half-rectangle room suggests continuing exploration into the unexplored side. Human beings would naturally pay attention to both local observations and the global map.
    \item Avoiding meaningless stuck. Due to limited egocentric views and VLM reasoning errors, previous methods often get stuck in corners—even when actions like "turn around" are explicitly provided, as seen in \cite{vlmnav,navvlm}. With global perception, such failures are rare, as it enables the agent to look beyond local surroundings and move efficiently toward the goal.
    % \item Achieving effective and fast system. Previous works often access VLM multiple times at each navigation step for next move, and move the robot a short distance, while with global perception, the access to VLM reduced and distance of move go up.
\end{itemize}
Following these ideas, our methods organically combined these two perceptions together.

\subsection{Local perception}
Local perception refers to the RGB-D data captured by the robot’s egocentric sensor. It provides detailed information about the immediate surroundings, allowing the robot to detect and segment the target object from the image for precise localization.

After acquiring the RGB-D input, the RGB image is processed to detect the target object. The detection is provided by Qwen-VL model. If the object is found, we perform goal projection refinement (explained in \ref{refinement}). The refined region in the egocentric view is then projected onto the top-down map using depth information, serving as a target for path planning.

This projection is done using depth data and point cloud generated by a classic SLAM method. We use the depth image to compute a 3D point cloud of the current view, then voxelize it to represent occupancy. By selecting a specific height range, we generate a local top-down map, which is then merged with existing global top-down map. During this process, the refined target area from the egocentric view is also projected onto the top-down map as the navigation goal.

\subsection{Goal projection refinement}
\label{refinement}
When the local perception module detects the goal object, the Qwen-VL model provides a bounding box of the object in the observation. However, directly using bounding box projection to the top-down map has two main problems that possibly lead to navigation failure.

\begin{figure}
    \centering
    \includegraphics[width=1\linewidth]{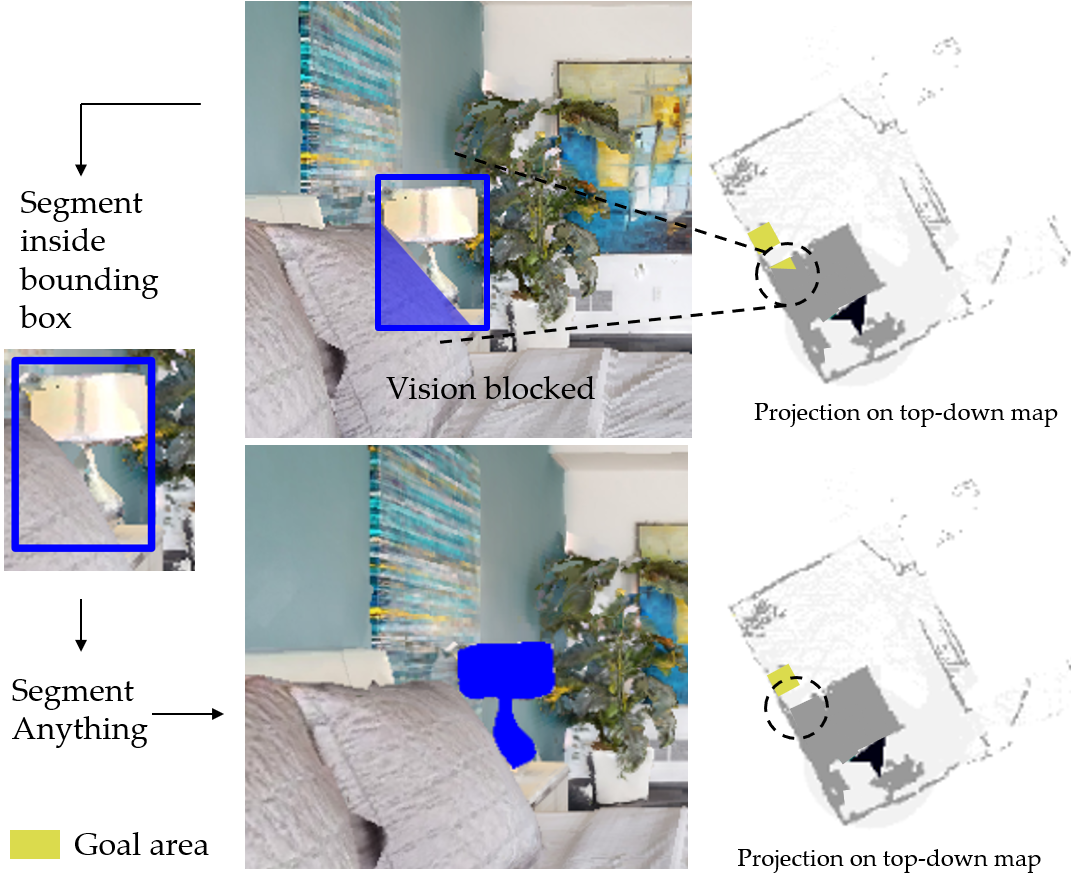}
    \caption{Refine egocentric goal area with segment anything. Taking ``lamp" as an example, directly using the detection bounding box can lead to inaccuracies and unexpected areas after projection onto a top-down map.}
    \label{fig:vision_block}
\end{figure}

First, as illustrated in Figure \ref{fig:vision_block}, as the object may be partially occluded from the robot's view, causing inaccuracy in goal area projection on top-down map. To overcome this, following \cite{grounded_sam}, we further applied \cite{mobile_sam} to segment the objects after Qwen-VL detection inside the bounding box to refine the goal object area in egocentric observation before the projection. We further erode the segmentation area on observation to exclude boundary regions that may introduce projection inaccuracies. This approach effectively mitigates the issue of visual occlusion. 

Second, if the goal object is spatially surrounded by other objects—such as a lamp on a nightstand—its projected area on the top-down map may be enclosed by non-goal obstacles. In such cases, standard A-star path planning may fail, as the target location becomes effectively unreachable. To resolve this, we apply morphological dilation—a classic computer vision technique—on the top-down map to expand the goal area beyond the surrounding obstacles, ensuring it remains accessible to the planner. 

This post-processing step improves HyPerNav and reduces many failure cases, as shown in Figure \ref{fig:goal_refinement}.

\subsection{Global perception}
The global perception refers to the rich information on the gradually constructed during the exploration to the navigation goal. Our aim is to use the VLM as a "brain" for global navigation—analyzing map clues and suggesting the next exploration location. To achieve this, following previous works\cite{vlmnav,navvlm}, we design a VQA interface for the VLM. 

The input to the VLM includes a contextual image showing the robot’s current position and orientation (as an arrow) on the top-down map, along with its past trajectory. We also divide the explored area into grid blocks, each labeled with a unique ID, overlaid on the map. Given a prompt such as ``To find \textless{}target object\textgreater{}, which block should you go" along with this contextual image, the VLM outputs a block number after a post-processing text summarizer. This block is then treated as the next exploration area. A post-processing step selects a free (non-obstacle) location within that block as the actual short-term destination.

To prevent the global perception module from repeatedly suggesting the same location, we maintain a visited memory in the world frame that records previously visited positions. If the proposed goal falls within the vicinity of a past location, we add an explicit instruction like ``don't answer number [x,...]" to the prompt and ask the VLM to generate a new block number.

The robot will not receive a new global short-term destination until a ``destination update” flag becomes true. This flag is set to true under the following conditions: 1. The current short-term goal is reached, which is the most common case; 2. A human-set endurance limit is exceeded, which is used to handle unexpected issues, such as drifting out of bounds due to incomplete depth data in simulation, or getting stuck moving in one direction; 3. The path planning module fails to find a valid path to the proposed short-term goal. Once any of the above conditions is satisfied, the flag is then set back to false.

Both local and global perception can provide destination areas. However, global perception offers potential exploration directions, while local perception provides precise goal for termination. Therefore, local perception takes priority over global perception during navigation. 

% Path to destination position from both perceptions is computed using the same algorithm (A star) and updated through the same mechanism.

\subsection{Path planning and moving}
From the robot’s position to the next destination, we dynamically search and update the path using A star search as the top-down map gradually becomes more complete. The path planning module re-computes the path every 10 steps or whenever the current path is blocked by newly discovered obstacles, ensuring smooth robot movement.

To avoid collisions, we apply morphological dilation to the obstacle map. For each A star search, the point within projection area and closest to the robot’s current position is selected as the destination point, and the robot’s position is used as the start point. If either the start or destination point lies inside an obstacle, it is replaced with the nearest free (non-obstacle) point before path computation.

The path generated by the A star algorithm consists of a sequence of fine-grained waypoints, and the robot follows this path by combining primitive actions: turning left, turning right, and moving forward. This module calculates the relative angle between the robot’s current orientation and the direction to the next waypoint. If the angle exceeds the turning threshold of 15° to either the left or right, the robot executes the corresponding turn action; otherwise, it moves forward. Each turn action adjusts the robot’s orientation by 30°.

\subsection{Termination}
Knowing when to terminate is important to avoid stopping too far from the target or passing it without stopping. Ideally, navigation terminates when the robot reaches the goal provided by local perception, which is often the case in experiments. The distance threshold for reaching the goal is set manually. As a double-check, we query the VLM with a prompt to determine whether the target object is present. Finally, a maximum step limit serves as a fallback termination condition, though reaching this limit typically means the navigation has failed.

\section{Simulation Evaluation}
We evaluate HyPerNav on two types of object navigation tasks: the traditional vocabulary-based navigation dataset Habitat-Matterport3D (HM3D) \cite{hm3d}, and the open-vocabulary navigation dataset Open-Vocabulary Object Goal Navigation (OVON) \cite{ovon}.

\begin{figure}
    \centering
    \includegraphics[width=0.9\linewidth]{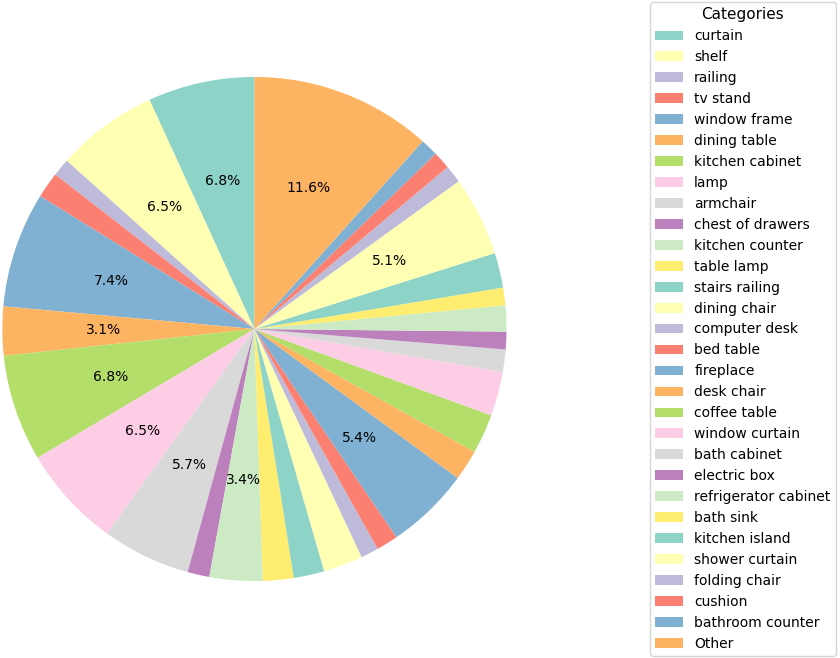}
    \caption{Goal object category frequency distribution in OVON}
    \label{fig:ovon_distribution}
\end{figure}

Both datasets are based on Matterport3D (MP3D) \cite{mp3d} scenes but feature different types of navigation goals. HM3D primarily includes in-domain object categories such as ``bed" and ``plant", while OVON contains more diverse and challenging object names, including compositional phrases like ``L-shaped sofa" and ``clothes dryer". The overall difficulty of OVON is higher than that of HM3D. Each scene typically contains around 100 episodes in HM3D. We use the ``validation unseen" split of OVON, which consists of 3,000 episodes in total, with an average of 83 episodes per scene. The distribution of target objects in this split is shown in Figure \ref{fig:ovon_distribution}.

We explicitly report the scene IDs used in our ObjNav evaluation, and we encourage future work to follow this practice to improve reproducibility and comparability.

The performance metrics for navigation are Success Rate (SR) and Success weighted by Path Length (SPL) \cite{embodied_eval}. SR measures the proportion of episodes in which the robot successfully reaches the goal object. SPL measures the efficiency of the navigation, reflecting how closely the robot’s trajectory approaches the optimal geodesic shortest path. For the goal projection refinement, we applied Qwen-VL detection and MobileSAM\cite{grounded_sam} segmentation. For goal area morphological dilation, we applied 5x5 kernel and 3 iteration via OpenCV. We implemented the experiments via habitat\cite{habi3} on GTX4080 server. The atom actions of robot are ``forward 0.5m", ``turn left 30°" and ``turn right 30°". We set maximum step as 500 steps. We used QwenVL-7B as main VLM. Note all the numbers in the table below are percentage.

\subsection{Experiment results}
In Table \ref{tab:performance_avg}, the baselines' data are from both training and training-free popular navigation works which have documented performance statistics on HM3D. A more detailed performance is in Table \ref{tab:performance_hm3d} and Table \ref{tab:performance_ovon}. However, due to length, Table \ref{tab:performance_hm3d} includes 1,000 navigation episodes, and Table \ref{tab:performance_ovon} includes 961 navigation episodes.

\begin{table}[h]
    \centering
    \caption{Average Performance in HM3D, OVON}
    \begin{tabular}{|c|c|c|c|c|c|}
        \hline
        Method & Training- & \multicolumn{2}{c|}{HM3D} & \multicolumn{2}{c|}{OVON} \\
        \cline{3-6}
        & free & SR↑ & SPL↑ & SR↑ & SPL↑ \\
        \hline
        PIRLNav\cite{pirlnav}& No&\textbf{64.1}& 27.1 & - & - \\
        \hline
        ZSON\cite{majumdar2022zson}& No&25.5&12.6 &- &- \\
        \hline
        VLFM\cite{vlfm}& Yes&52.5&30.4 &- &- \\
        \hline
        NavVLM\cite{navvlm}& Yes&48.0&33.5 &- &- \\
        \hline
        PIVOT\cite{pivot}& No&24.6&10.6 &- &- \\
        \hline
        VLMNav\cite{vlmnav}& Yes&50.4&21.0 &- &- \\
        \hline
        DyNaVLM\cite{ji2025dynavlmzeroshotvisionlanguagenavigation}& Yes&45.0&23.2 &- &- \\
        \hline
        Ours& Yes &\textbf{53.4} & \textbf{35.8} &65.4 &43.7 \\
        \hline
    \end{tabular}
    
    \label{tab:performance_avg}
\end{table}

\subsection{Experiment analysis}
In Table~\ref{tab:performance_avg}, HyPerNav achieves the highest SPL among all baselines on HM3D, with a second-highest success rate.

\textbf{HyPerNav performs effectively in ObjNav}. In terms of SR, some baselines such as VLFM~\cite{vlfm} use step limits close to 50,000, allowing extensive exploration. In contrast, efficient full-space exploration in HM3D typically requires only 500–3,000 steps~\cite{frontiernet}. HyPerNav, however, operates under a strict limit of 500 steps and still achieves the second-highest success rate. While PIRLNav reports a slightly higher SR, training-based methods generally face challenges in scene data generation and generalization across diverse environments.

\textbf{HyPerNav generates more efficient navigation paths}. In terms of SPL, HyPerNav produces shorter and more direct trajectories toward the target compared to both training and training-free baselines, even when its SR is lower. This suggests that HyPerNav tends to follow paths closer to the geodesic shortest route, which aligns with expectations for practical robot navigation.

\textbf{HyPerNav handles complex language goals}. We also evaluate HyPerNav on the more challenging OVON dataset, where target objects are described using detailed phrases. Despite the fact that initial robot positions in OVON are usually closer to the target than in HM3D, the diversity and complexity of object categories increase the demand on language understanding. HyPerNav is able to interpret such descriptions correctly in many cases, though it occasionally reaches semantically similar but incorrect objects. For example, navigating to a ``window curtain" instead of a "shower curtain". A detailed failure analysis on OVON is provided in Figure~\ref{fig:failure_stat}(b).

\begin{table}[htbp]
\centering
\caption{clip of HyPerNav detailed performance in HM3D}
\begin{tabular}{|c|c|c|c|c|c|}
\hline
ID & SR↑ & SPL↑ & ID & SR↑ & SPL↑ \\
\hline
802 & 47.5 & 33.1 & 813 & 61.5 & 40.3 \\
\hline
814 & 37.5 & 22.6 & 824 & 55.6 & 39.5 \\
\hline
829 & 51.9 & 36.9 & 835 & 47.5 & 40.8\\
\hline
853 & 62.6 & 42.4 & 876 & 41.4 & 27.7\\
\hline
877 & 59.3 & 33.0 & 880 & 59.6 & 31.8\\
\hline
\end{tabular}

\label{tab:performance_hm3d}
\end{table}

\begin{table}[htbp]
\centering
\caption{clip of HyPerNav detailed performance in OVON}
\begin{tabular}{|c|c|c|c|c|c|}
\hline
ID & SR↑ & SPL↑ & ID & SR↑ & SPL↑ \\
\hline
820 & 75.0 & 50.7 & 835 & 68.7 & 46.4  \\
\hline
810 & 50.0& 38.2 & 847 & 76.9 & 52.2 \\
\hline
821 & 38.9 & 29.8 &831 & 42.9 & 22.8  \\
\hline
853 & 70.2 & 42.6 & 873 & 75.9 & 59.2 \\
\hline
877 & 67.3 & 45.9 & 878 & 80.3 & 55.2 \\
\hline
880 & 86.7 & 45.7 & 890 & 63.6 & 44.2 \\
\hline
\end{tabular}

\label{tab:performance_ovon}
\end{table}

\begin{figure*}
    \centering
    \includegraphics[width=1\linewidth]{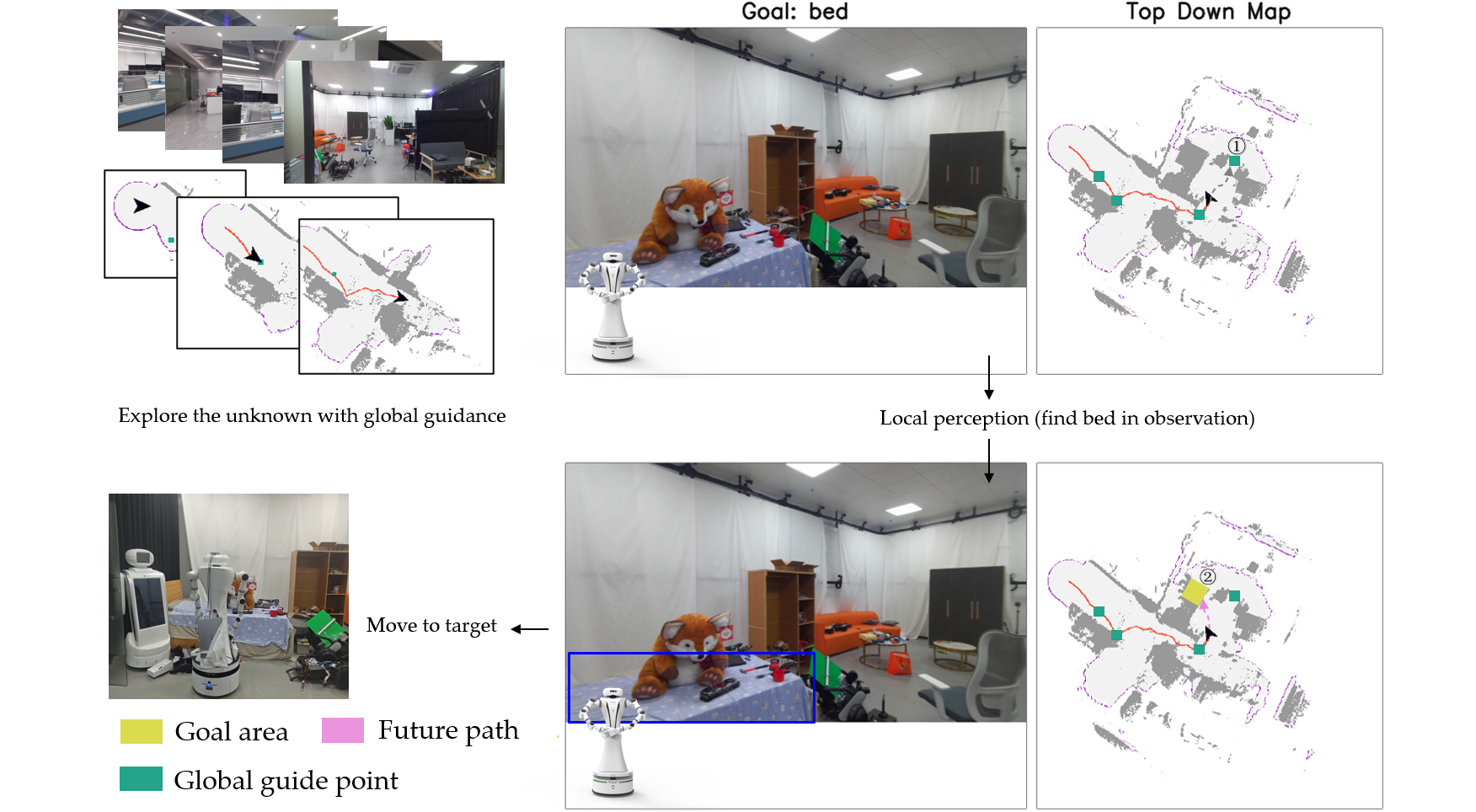}
    \caption{In the rightside illustration, the robot would follow global perception guide along path 1 (grey path) if local perception does not find the target bed. As target object is detected by local perception, the robot will stop following path 1 and start following local perception path 2 (pink path) until reaching the object. The ObjNav is then terminated.}
    \label{fig:validation2}
\end{figure*}

\subsection{Ablation study}
We conducted the ablation study in a single scene, as running it across all scenes would be time-consuming.

\begin{figure}[t]
\centering

\begin{tabular}{c}
\includegraphics[width=0.8\linewidth]{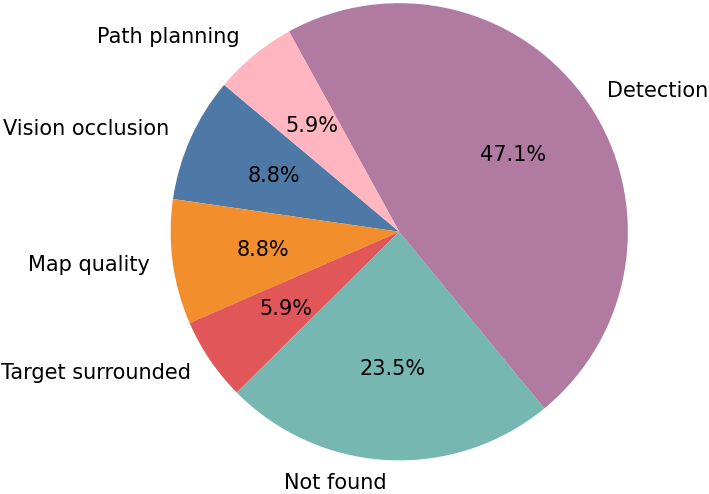} \\
(a) \\[1em]
\includegraphics[width=0.9\linewidth]{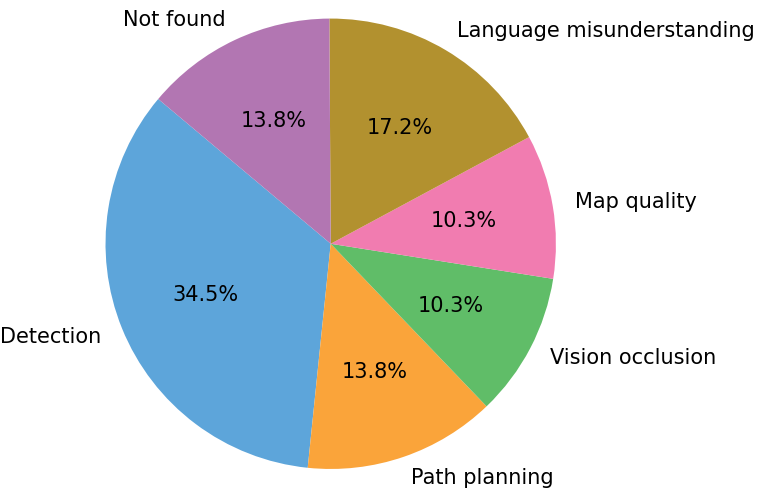} \\
(b)
\end{tabular}

\caption{(a) Failures statistics of HM3D. (b) Failures statistics of OVON.}
\label{fig:failure_stat}
\end{figure}

\textbf{What contributes to the navigation failures?} 
We analyze failure cases on HM3D and OVON. As shown in Figure~\ref{fig:failure_stat}(a), ``Detection" failures arise from false positives due to poor viewpoints or detection limits. ``Map quality" refers to the absence of RGB-D  data in simulation leading to wrong path planning. ``Not found" occurs when the robot fails to reach or observe the target within 500 steps, usually due to ineffective exploration guidance—this is the main failure mode, indicating room for improvement. ``target surrounded" refers to target area is enclosed by obstacles thus path planning fails, as the simple dilation mitigates but not solves this problem. ``path planning" is when plan planning generates a wrong path or does not update the path correctly, usually because of inaccuracy in robot position and goal area. In OVON, many failures stem from "Language understanding", where the robot finds a semantically similar object (e.g., ``window curtain") instead of the correct one (e.g., ``shower curtain").
% \begin{figure}
%     \centering
%     \includegraphics[width=1\linewidth]{failures.png}
%     \caption{Failure factors}
%     \label{fig:failure_stat}
% \end{figure}

\begin{figure}
    \centering
    \includegraphics[width=1\linewidth]{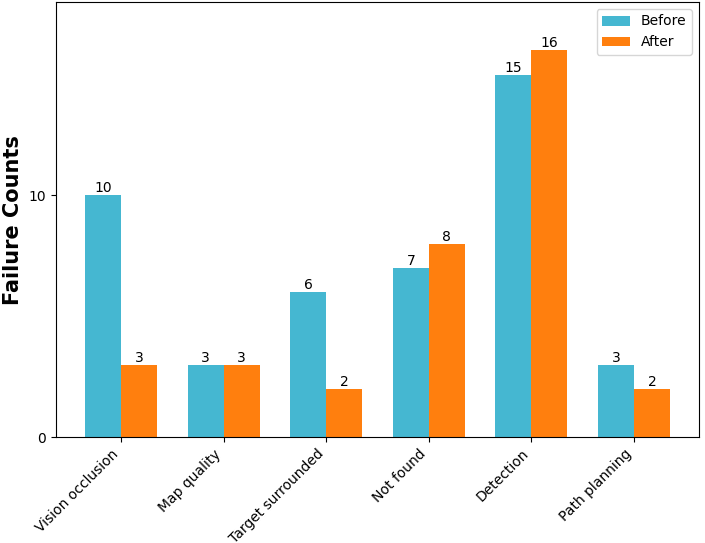}
    \caption{Comparison of goal projection refinement}
    \label{fig:goal_refinement}
\end{figure}

\textbf{How much does goal refinement helps the navigation?} As illustrates in Figure \ref{fig:goal_refinement}, vision occlusion and goal in obstacle are greatly mitigated by goal projection refinement, which drops 70\% and 67\% respectively.

\textbf{What is the HyPerNav navigation time consumption?} HyPerNav is relatively time-effective. Compared with other VLM-based framework~\cite{vlmnav,navvlm} to guide navigation in egocentric observation, HyPerNav does not necessarily require VLM language description of the scene at every step, reducing the time consumed by VLM long sentence inference. Compared with \cite{navvlm}, time consumption of each step is 3.3 seconds while HyPerNav's is 1.2 second.

\section{Real-world Validation}
We conduct real-world navigation validation on a half-human with a wheel base robot. We used Microsoft Azure DK RGB-D sensor on the robot and get the point cloud directly from the equipment API. We used IMU-based odometry to obtain robot position and orientation. The hyper-parameters such as forward meters and voxel clipping are slightly different but the method scheme are identical. We implemented VLM serving on a RTX 4090 server and used self-developed HTTP Flask server API to communicate between the robot sensor and VLM reasoning.

Our real-world experiment environment is our lab office. We choose two object categories ``bed" and ``umbrella" as navigation targets. We illustrate a detailed episode on ``bed" in Figure \ref{fig:validation2}, and report the overall experiment performance in Table \ref{tab:validation}. We further compared HyPerNav to frontier\cite{frontier} strategy. Both methods used the same detection and segmentation module. We recorded their average successful navigation path length via accumulated position coordinates provided by odometry, and manually measured the optimal geodesic path.

\begin{table}[t]
\centering
\begin{tabular}{|c|c|c|}
\hline
bed    & frontier & HyPerNav\\ \hline
Succ/All & 3/5 & 4/5 \\ \hline
SR       & 60\% & 80\% \\ \hline
SPL     & 0.24 & 0.5 \\ \hline
umbrella    & frontier & HyPerNav \\ \hline
Succ/All & 4/5 & 3/5 \\ \hline
SR       & 80\% & 60\% \\ \hline
SPL     & 0.35 & 0.42 \\ \hline

\end{tabular}
\caption{Real-world ablation on goal ``bed" and ``umbrella".}
\label{tab:validation}
\end{table}

In Table \ref{tab:validation}, Frontier achieved a high success rate, which was anticipated since both target objects fall within the detection vocabulary and the frontier-nearest exploration strategy ensures thorough exploration of the lab space. However, HyPerNav matches this success rate while achieving higher SPL in more efficient exploration. This is because global perception enables HyPerNav to capture obstacle shape information, providing additional hints beyond what the frontier-based approach can offer.

HyPerNav achieves a higher success rate in navigating to a "bed" compared to an "umbrella," due to the bed's location being more predictable and its shape being more easily recognizable on a top-down spatial map. In contrast, smaller objects like an "umbrella" present greater challenges. These items are not only less predictable in terms of placement but also tend to have less distinctive shapes from a top-down perspective, making them harder to identify.

HyPerNav has better SPL on both goal objects, demonstrating the effectiveness of ObjNav ability. By leveraging information from both the global top-down map and local observations, HyPerNav makes better use of environmental cues, enhancing navigation efficiency.

\section{Conclusion}
In this paper, we proposed HyPerNav, a training-free approach for effective Object Navigation (ObjNav), designed to address the disjointed perception of local and global sources. HyPerNav leverages hybrid perceptions by integrating egocentric observations and  top-down map observation, utilizing VLMs to extract cues from both local and global contexts. We conducted extensive simulations on two ObjNav datasets: standard object categories in HM3D and complex language objects in OVON. The results demonstrate that HyPerNav achieves effective navigation performance in both settings. We further present ablation studies on failure factor analysis, and goal refinement modules. Additionally, real-world validation and comparison was carried out using a half-human robot in our laboratory environment, confirming the effectiveness and navigational intelligence of HyPerNav. 


%%%%%%%%%%%%%%%%%%%%%%%%%%%%%%%%%%%%%%%%%%%%%%%%%%%%%%%%%%%%%%%%%%%%%%%%%%%%%%%%

%%%%%%%%%%%%%%%%%%%%%%%%%%%%%%%%%%%%%%%%%%%%%%%%%%%%%%%%%%%%%%%%%%%%%%%%%%%%%%%%

%%%%%%%%%%%%%%%%%%%%%%%%%%%%%%%%%%%%%%%%%%%%%%%%%%%%%%%%%%%%%%%%%%%%%%%%%%%%%%%%

\section*{multimedia}
Explore the real-world robot validation videos available in our multimedia materials to gain deeper insights into HyPerNav practical applications.
\begin{figure}[htbp]
    \centering
    \includegraphics[width=0.6\linewidth]{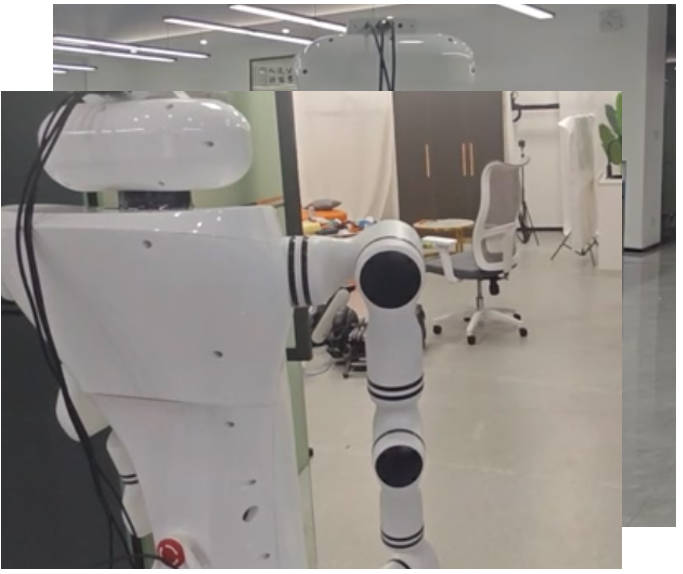}
    \caption{Real-world robot episode is available in the multimedia materials.}
    \label{fig:placeholder}
\end{figure}

% This work was supported by NSFC with Grant No. 62293482, by the Basic Research Project No. HZQB-KCZYZ-2021067 of Hetao Shenzhen HK S\&T Cooperation Zone, by Shenzhen General Program No. JCYJ20220530143600001, by Shenzhen-Hong Kong Joint Funding No. SGDX20211123112401002, by the National Key R\&D Program of China with grant No.2018YFB1800800, by the Shenzhen Outstanding Talents Training Fund 202002, by Guangdong Research Project No. 2017ZT07X152 and No. 2019CX01X104, by the Guangdong Provincial Key Laboratory of Future Networks of Intelligence (Grant No. 2022B1212010001), by the Guangdong Provincial Key Laboratory of Big Data Computing, The Chinese University of Hong Kong, Shenzhen, by the NSFC 61931024\&12326610, by the Shenzhen Key Laboratory of Big Data and Artificial Intelligence (Grant No. ZDSYS201707251409055), and the Key Area R\&D Program of
% Guangdong Province with grant No. 2018B03033800, by Tencent\&Huawei Open Fund. Zecheng Yin is currently with the Shenzhen Future Network of Intelligence Institute (FNii-Shenzhen), and the Guangdong Provincial Key Laboratory of Future Networks of Intelligence, the Chinese University of Hong Kong at Shenzhen.

\bibliographystyle{IEEEtranS}
\bibliography{IEEEexample}

\end{document}